\documentclass{article}
\usepackage{iclr2027_conference,times}

\usepackage{amsmath,amsfonts,bm}

\def\eqref#1{equation~\ref{#1}}
\def\1{\bm{1}}

\DeclareMathAlphabet{\mathsfit}{\encodingdefault}{\sfdefault}{m}{sl}
\SetMathAlphabet{\mathsfit}{bold}{\encodingdefault}{\sfdefault}{bx}{n}

\usepackage{amsmath,amssymb,amsfonts,mathtools,amsthm}
\usepackage{graphicx}
\usepackage{wrapfig}
\usepackage{subcaption}
\usepackage{booktabs}
\usepackage{multirow}
\usepackage{nicefrac}
\usepackage{microtype}
\usepackage[utf8]{inputenc}
\usepackage{tabularx}
\usepackage{enumitem}
\usepackage{algorithm}
\usepackage{algorithmic}
\usepackage{colortbl}
\usepackage{longtable}
\usepackage{url}
\usepackage{tikz}
\usetikzlibrary{arrows.meta,fit,backgrounds}
\usepackage{hyperref}
\usepackage[capitalize,noabbrev]{cleveref}
\definecolor{resultrow}{RGB}{232,240,248}
\definecolor{groupgray}{gray}{0.94}
\definecolor{visblue}{RGB}{223,236,250}
\definecolor{langtan}{RGB}{253,233,217}
\definecolor{coregray}{RGB}{237,237,240}
\definecolor{actgreen}{RGB}{225,243,229}
\newcommand{\method}{\textsc{SimVLA}\,2.0}
\newcommand{\rev}[1]{#1}

\title{SimVLA 2.0: Instruction-Conditioned Control without Language Pretraining}
\author{%
  Yuankai Luo\textsuperscript{2,3,1} \quad Woping Chen\textsuperscript{1} \quad
  Tong Liang\textsuperscript{1} \quad Zhenguo Li\textsuperscript{1}\thanks{Corresponding author.} \\
  \textsuperscript{1}Frontier Robotics \\
  \textsuperscript{2}State Key Laboratory of Novel Software Technology, Nanjing University \\
  \textsuperscript{3}School of Artificial Intelligence, Nanjing University \\[0.2em]
  \small Project website: \url{https://frontierrobo.github.io/SimVLA2}%
}
\hypersetup{
  hidelinks,
  pdftitle={SimVLA 2.0: Instruction-Conditioned Control without Language Pretraining},
  pdfauthor={Yuankai Luo, Woping Chen, Tong Liang, Zhenguo Li}
}
\iclrfinalcopy

\begin{document}
\maketitle
% Use the signed title block without the template's conference publication header.
\fancyhead[L]{}

\begin{abstract}
Vision-Language-Action policies typically retain pretrained language components, yet on finite, recurring task inventories the role of that retention is rarely tested directly.
Action training itself can erode the competence a pretrained backbone contributes, and what remains often fails to transfer under meaning-preserving paraphrases.
We therefore build the opposite end of the design space: instructions remain, but no pretrained language-side component is loaded.
\method{} loads no language model, text encoder, or token embedding: a fixed, parameter-free mapping indexes randomly initialized embeddings learned solely from action supervision, and a joint action Transformer conditions a continuous action head on the embeddings and the pretrained vision encoder's features.
Across seven benchmark evaluations spanning six manipulation domains, it remains within the range of recent language-pretrained systems---LIBERO 96.4\%, LIBERO-PRO 96.5\% under semantic instruction perturbations, CALVIN ABC$\rightarrow$D a 4.24 mean chain length, WidowX 91.7\%, Google Robot 82.0\%, and RoboTwin 2.0 Easy/Hard 94.0\%/87.2\%.
\rev{On the BEHAVIOR-1K Public Validation split, \method{} achieves a Q-Score of \(0.2450\).}
Ablations separate the sources of performance: instruction conditioning is essential, pretrained token embeddings provide no consistent advantage over action-trained codes, and strong visual pretraining remains important. 
% Code is available at \url{https://github.com/LUOyk1999/SimVLA2}.
\end{abstract}

\section{Introduction}
\label{sec:introduction}

Vision-Language-Action (VLA) research increasingly couples pretrained multimodal backbones with autoregressive, diffusion, or flow-matching action generators~\citep{kim2024openvla,pi0,pi05}.
This recipe is effective, but it leaves a basic attribution question unresolved.
When a benchmark repeatedly draws instructions from a finite task inventory, does control require pretrained linguistic knowledge, or can the instruction act as a stable task code whose meaning is acquired from demonstrations?
We approach this empirically rather than analytically: we construct a policy that is conditioned on instructions but uses no pretrained language components, and measure how far it gets.

\paragraph{Why the necessity question is open.}
The usual defense of the language-model component is linguistic generalization. A pretrained backbone should let a policy follow ``pick up the banana'', ``grasp the banana'', and ``get the yellow fruit'' interchangeably.
Recent diagnostics arrive from two directions.
First, the inherited competence does not survive action training: \citet{UAM} report that standard end-to-end VLA fine-tuning removes almost all of the backbone's multimodal understanding, and that vision--language co-training reduces but does not eliminate this ``embodiment tax''.
Second, whatever language ability remains does not produce paraphrase generalization.
Under meaning-preserving rewordings of LIBERO instructions, seven VLA configurations spanning 0.6B--7.5B and four architecture families lose 22.8--51.9 points of success, and 80--96\% of those failures consist of wrong-task selection rather than imprecise execution~\citep{LIBEROPara}.
The degradation is driven mainly by substituting object words, which indicates surface-form matching rather than semantic grounding, and quadrupling task-level data diversity leaves it essentially unchanged.
\citet{GSR} observe the same pattern with drops as large as 67.5 points.
The pattern extends to the strongest systems in our comparison tables: under LIBERO-Para's paraphrase protocol~\citep{LIBEROPara}, X-VLA falls from 97.8 to 62.1 and VLA-Adapter from 98.2 to 46.3.

The two findings converge.
If the pretrained language component is substantially degraded by action training, and if what remains behaves like surface-form task matching on the very benchmarks used to validate VLAs, then it is not established that language pretraining is the ingredient producing instruction-conditioned control success.
We do not claim this ceiling is permanent; \citet{UAM} and \citet{GSR} both propose remedies, and the mechanism is still debated, with the latter arguing that task semantics are retained internally but not routed usefully to the action head.
Our point is narrower: on the task-bounded benchmarks used to validate VLAs, the assumed benefit is realized far less than the architecture implies.
That makes the opposite end of the design space worth measuring.
How far a policy that keeps instructions but uses no pretrained language components can go is the question this paper answers.

\method{} makes that intervention explicit.
\textbf{\method{} uses no language model.}
Throughout this paper, ``language pretraining'' operationally means loading (i) a pretrained language model, (ii) a pretrained text encoder, or (iii) pretrained token embeddings; \method{} loads none of the three. The vision encoder's image--text pretraining is retained by default and ablated separately.
We call this construction language-model-free (LM-free).
It loads only a pretrained vision encoder. A parameter-free mapping assigns each instruction to deterministic fixed codes, and the corresponding embedding table is randomly initialized and updated solely by action-supervised trajectories; the mapping can be replaced by a fixed random-symbol relabeling. 
The resulting vectors are contextualized with visual patches and proprioception inside a joint action Transformer; they are not pretrained sentence representations.

The published SimVLA baseline controlled data order, normalization, optimization, horizon, and backbone adaptation before attributing gains to architecture on the benchmarks it studied, and retained a pretrained vision--language backbone throughout~\citep{luo2026simvla}.
We apply that controlled-recipe discipline to the component that baseline retained as pretrained---the language side of its vision--language backbone---removing pretrained language components while keeping the instruction input.
The policy is still a vision-language-action model in the interface sense: it is conditioned on natural-language instructions, but it drops the pretrained language model rather than the language input.

Removing language pretraining from the action policy does not imply that language is unimportant: a full VLM remains essential for open-ended instruction understanding, task decomposition, planning, and reasoning.
We view \method{} as a bounded instruction-conditioned controller that can operate beneath such a high-level model, rather than as a replacement for its linguistic capabilities.

\begin{table}[t]
  \centering
  \vspace{-0.2 in}
  \caption{\textbf{LM-free architecture and headline results.}
  The left panel abstracts the conditioning path: the vision branch loads a pretrained encoder, whereas the instruction branch (dashed) loads no pretrained weights and is \rev{a fixed tokenizer with token embeddings learned from action supervision.}
  The action head is drawn as a generic interface because the intervention is orthogonal to the action objective; \Cref{sec:method} specifies the instantiation used in all reported results.
  The right panel summarizes \method{} results; benchmark-specific protocols appear in \Cref{sec:evaluation}.
  \rev{For BEHAVIOR-1K, the rank compares the Public Validation Q-Score with the archived Standard-track scores.}}
    % ``LIBERO-PRO Sem.'' is the mean success under semantic instruction perturbations.
  \label{tab:overview}
  \begin{minipage}[t]{0.56\linewidth}
  \vspace{0pt}
  \centering
  \begin{tikzpicture}[
    baseline=(current bounding box.north),
    font=\scriptsize,
    >={Latex[length=1.3mm,width=1.1mm]},
    bx/.style={draw=black!55, rounded corners=1.5pt, align=center, inner sep=2.2pt, text width=2.75cm},
    vis/.style={bx, fill=visblue},
    lang/.style={bx, fill=langtan, draw=orange!70!black},
    wide/.style={draw=black!55, rounded corners=1.5pt, align=center, inner sep=2.5pt, text width=6.1cm, fill=coregray},
    act/.style={wide, fill=actgreen},
    fl/.style={->, semithick, black!60}
  ]
  \node[vis]  (img) at (-1.62,0)     {observation};
  \node[lang] (ins) at ( 1.62,0)     {instruction};
  \node[vis]  (enc) at (-1.62,-1.0)  {\textbf{vision encoder}\\\emph{pretrained}};
  \node[lang] (emb) at ( 1.62,-1.0)
    {\textbf{instruction embedding}\\
     \emph{randomly initialized}\\[-0.5pt]
     {\color{orange!70!black}\textbf{LM parameters: 0}}};
  \node[wide] (head) at (0,-2.62)
    {\textbf{action head}\\[-1.5pt]
     {\tiny flow matching \textbar{} diffusion \textbar{} autoregressive \textbar{} regression}};
  \node[act]  (out) at (0,-3.62) {action chunk};
  \begin{scope}[on background layer]
    \node[draw=orange!70!black, dashed, rounded corners=3pt, inner sep=3.5pt, fit=(ins)(emb)] {};
  \end{scope}
  \draw[fl] (img) -- (enc);
  \draw[fl] (ins) -- (emb);
  \draw[semithick, black!60] (enc.south) |- (0,-1.92);
  \draw[semithick, black!60] (emb.south) |- (0,-1.92);
  \draw[fl] (0,-1.92) -- (head.north)
    node[midway, right=1.5pt, font=\scriptsize] {$+$ proprioceptive state};
  \draw[fl] (head) -- (out);
  \end{tikzpicture}
  \end{minipage}\hfill
  \begin{minipage}[t]{0.42\linewidth}
  \vspace{0pt}
  \centering
  \scriptsize
  \vspace{0.5em}
  \setlength{\tabcolsep}{3pt}
  \resizebox{\linewidth}{!}{%
  \begin{tabular}{lc}
  \toprule
  \textbf{Evaluation} & \textbf{\method{}} \\

\multicolumn{2}{r}{\rev{0.3B vision backbone, 0.6B total}} \\
  \midrule
  LIBERO Avg. (\%) & 96.4 \\
  LIBERO-PRO Sem. (\%) & 96.5 \\
  CALVIN Avg. length & 4.24 \\
  WidowX Avg. (\%) & 91.7 \\
  Google VA Avg. (\%) & 82.0 \\
  RoboTwin Easy/Hard (\%) & 94.0/87.2 \\
  \rev{BEHAVIOR-1K Q-Score} & \rev{0.2450 (2nd)} \\
  \bottomrule
  \end{tabular}}
  \vspace{0.5em}

  \end{minipage}
  \vspace{-0.1 in}
  \end{table}

\paragraph{What the scores establish.}
Our central finding is a sufficiency result: across tabletop manipulation, controlled distribution shifts, long-horizon chains, real-to-sim evaluation on two embodiments, bimanual control, and household behavior, removing every pretrained language parameter still leaves a policy that scores in the range of recent language-pretrained systems (\Cref{tab:overview}).
Ablations separate the sources of performance: instruction conditioning is essential, \rev{the evaluated pretrained text components offer no consistent advantage over action-trained codes}, and strong visual pretraining remains important.

The instruction therefore carries the task signal. \rev{The policy tolerates paraphrases in LIBERO-PRO and LIBERO-Para and executes an unseen instruction string composed of training-seen tokens on WidowX.}

\noindent\textbf{Contributions.}
\begin{itemize}[leftmargin=*,nosep]
    \item We construct the missing design point described above: \method{} receives instructions but loads no pretrained language-side parameters and learns its instruction embeddings from action supervision alone.
    \item Across seven benchmark evaluations spanning six manipulation domains, \method{} reaches the headline values in \Cref{tab:overview}, placing an LM-free controller in the range of recent language-pretrained systems.
    \item Controlled ablations show that instruction conditioning is essential, pretrained token embeddings provide no consistent gain, and strong visual pretraining matters. Robustness evaluations \rev{characterize tolerance to rewording and how it varies with training-token coverage.}
\end{itemize}

\noindent\Cref{sec:related} situates this construction among existing systems, \Cref{sec:method} formalizes it, and \Cref{sec:evaluation} reports the evaluation results and ablations.

\section{Related Work}
\label{sec:related}

\paragraph{VLA backbones and action generation.}
RT-1 introduced tokenized robot control~\citep{brohan2022rt}; RT-2 showed that co-fine-tuning a web-scale vision-language model yields control policies that inherit web knowledge~\citep{brohan2023rt2}; OpenVLA made the recipe open-source~\citep{kim2024openvla}.
How actions are decoded is a largely orthogonal axis: diffusion-based action heads descend from Diffusion Policy~\citep{DiffusionPolicy}; generative and tokenized action interfaces are explored in $\pi_0$, $\pi_{0.5}$, and DD-VLA~\citep{pi0,pi05,DD_VLA}; and lightweight or modified decoders appear in OpenVLA-OFT, FAST, X-VLA, and VLA-Adapter~\citep{OpenVLA_OFT,FAST,x_VLA,VLA_Adapter}.
The published SimVLA baseline controlled data order, normalization, optimization, horizon, and backbone adaptation before attributing gains to architecture~\citep{luo2026simvla}; \method{} builds on that controlled-recipe discipline while removing pretrained language components.

\paragraph{Efficient and no-LLM execution.}
Compact policies such as SmolVLA, FLOWER, and VLA-Adapter reduce model or action-head cost~\citep{SmolVLA,FLOWER,VLA_Adapter}.
TurboVLA is a prominent efficiency-oriented system: it removes the large generative model but retains a compact pretrained text encoder~\citep{xie2026turbovla}, and VLA-0 removes large-scale action pretraining while keeping the pretrained VLM~\citep{goyal2025vla0}.
Neither removes the language-side pretraining that we study, and \method{} makes no matched latency or memory claim: its variable is language-side pretraining, not parameter count or runtime.

\paragraph{World and video-action models.}
A parallel family centers pretraining on explicit future video/action prediction; recent surveys organize these systems as world action models (WAMs) or video-action (VA) models~\citep{shen2026wam}, with representatives such as Motus, Fast-WAM, LingBot-VA, and WorldVLA~\citep{Motus,FastWAM,LingBotVA,WorldVLA}, and the family is still expanding~\citep{zhang2026nativevideo}.
Such systems still typically inherit a language or vision-language backbone, so they do not answer our question; our subtraction is complementary---we keep the action interface fixed and ask whether pretrained language components are needed at all.

\paragraph{Task representations and language conditioning.}
Task IDs, learned language interfaces, and pretrained sentence representations are long-standing conditioners; classic language-conditioned imitation learning suggests pretrained language helps synonym and held-out-task transfer~\citep{lynch2021language,jang2022bcz,mees2022what}. \rev{LIBERO compares language encoders with BERT-encoded task IDs~\citep{liu2023libero}, and \citet{mees2022what} study language encoder choices on CALVIN.}
Recipe and diagnostic studies complicate this picture: the RoboVLMs study systematically varies VLA recipes~\citep{li2024towards}, while diagnostic evidence shows that action fine-tuning erodes multimodal competence~\citep{UAM} and that paraphrase robustness collapses across architectures and scales~\citep{LIBEROPara,GSR}.
LIBERO-PRO complements these diagnostics with controlled object, position, semantic-instruction, and task perturbations of the standard LIBERO suites~\citep{zhou2025liberopro}.
These lines motivate our narrower question---whether fixed symbols suffice for task-bounded control---and differentiate us from recipe exploration: \method{} removes pretrained language-side parameters, while our ablations separately isolate instruction conditioning, pretrained token embeddings, and visual pretraining.

\paragraph{Cross-domain control.}
Generalist policies such as Octo and Open X-Embodiment target transfer across datasets and embodiments~\citep{Octo,o2024open}, while the protocols here span real-to-sim manipulation (SimplerEnv~\citep{li2024evaluating}), bimanual control (RoboTwin 2.0~\citep{chen2025robotwin}), and long-horizon household activity (BEHAVIOR-1K~\citep{BEHAVIOR1K}).
We evaluate across these complementary settings under their respective standard protocols.

\section{\texorpdfstring{\method{}}{SimVLA 2.0}}
\label{sec:method}

\subsection{Problem Formulation}
At control step \(t\), the policy receives valid RGB views \(I_t^{1:V}\), proprioceptive state \(s_t\), and instruction \(\ell\), and predicts
\[
A_t=[a_t,\ldots,a_{t+H-1}]\in\mathbb{R}^{H\times d_a}.
\]
At most \(h\leq H\) actions are executed before replanning.
Each benchmark uses its corresponding action adapter, normalization, \(H\), \(h\), and evaluator.
The shared goal is to model
\(\pi(A_t\mid I_t^{1:V},s_t,\ell)\)
without any pretrained language model, text encoder, or token embedding.

\subsection{Vision-Only Encoder and LM-Free Instruction Codes}
\paragraph{Vision encoder.}
Each view is encoded independently by a vision encoder \(E_\phi^{\mathrm{vis}}\),
\[
Z_t^v=E_\phi^{\mathrm{vis}}(I_t^{1:V}),
\]
which returns patch tokens.
We state this component abstractly: the construction requires only a vision-only backbone carrying no language-model component.
The intervention studied here concerns language pretraining rather than any particular visual backbone.

The reported runs instantiate \(E_\phi^{\mathrm{vis}}\) with pretrained SigLIP2-L/16 vision weights~\citep{tschannen2025siglip2}, retaining patch tokens and discarding the classification head.
We instantiate only the vision tower of SigLIP2 and never build its text tower, so SigLIP2 supplies visual pretraining without a language-model component.

\paragraph{Fixed instruction codes.}
A fixed tokenizer \(q\), with no trainable parameters, maps \(\ell\) to a sequence of discrete token IDs \(q(\ell)=(w_1,\ldots,w_L)\), where each \(w_i\) represents a word or WordPiece subword.
The corresponding instruction tokens are
\[
Z^\ell=[W_{w_1},\ldots,W_{w_L}],
\qquad
W\in\mathbb{R}^{|\mathcal V|\times d},
\]

where \(W\) is randomly initialized and updated exclusively by the action-supervised objective.
No text-only objective, pretrained token geometry, or dedicated contextual text encoder is used.
In the reported instantiation the lookup width \(d\) equals the vision encoder's output width (1024 for the SigLIP2-L runs), so no separate text projection is added at the embedding level.
A fixed random-symbol relabeling may permute code identities before training \rev{while preserving the WordPiece segmentation learned from text}.
Codes preserve discrete identity, sequence order, and shared symbols only; shared symbols across instructions are the only way different instructions share token geometry.
Thus \method{} removes language pretraining, not instruction input or subsequent multimodal contextual computation.

\subsection{Joint Action Transformer}
The trunk is orthogonal to the action objective; we describe the flow-matching instantiation used by all reported runs, whose auxiliary input is a noise level \(\tau\) and a corrupted chunk \(x_\tau\).
Each noisy action token \(x_{\tau,i}\), the proprioceptive state \(s_t\), and a sinusoidal time embedding \(e(\tau)\) are concatenated and mapped by a learned linear projection \(P_a\) to the trunk width \(d_{\mathrm{model}}\):
\[
h_i^a=P_a[x_{\tau,i};s_t;e(\tau)].
\]
Vision and instruction tokens are mapped by one shared learned linear projection \(P_{\mathrm{vlm}}\), so the encoder input is
\[
H_0=[h_1^a,\ldots,h_H^a;P_{\mathrm{vlm}}Z_t^v;P_{\mathrm{vlm}}Z^\ell],
\]
A learned positional embedding is added, and the sequence is processed by a standard bidirectional Transformer encoder~\citep{vaswani2017attention}.
A linear decoder on the first \(H\) positions predicts an action-space velocity.
In the reported runs, \(P_a\) maps the concatenation of action, proprioceptive, and time-embedding inputs to \(d_{\mathrm{model}}\), while \(P_{\mathrm{vlm}}\) maps the shared vision/lookup width \(d\) (1024) to the trunk width \(d_{\mathrm{model}}\) (1024 in all six training domains, \Cref{tab:hparams}).
Because instruction tokens attend jointly to vision, noisy actions, and \(\tau\), their hidden states are learned under action supervision and conditioned on noisy actions and flow time, rather than being reusable sentence embeddings.

\subsection{Action Head}
The components above produce a conditioning context
\[
c_t=g_\theta\!\left(I_t^{1:V},s_t,\ell\right),
\]
from which an action head \(\mathcal H_\theta\) decodes the chunk \(A_t\).
We state the head abstractly because the intervention studied in this paper acts on how \(c_t\) is formed, not on how \(A_t\) is decoded from it.
Let \(z\) denote whatever auxiliary input a head consumes, \(y\) the matching supervision target, and \(\ell_{\mathrm{obj}}\) the per-sample objective.
Training then minimizes
\[
\mathcal L(\theta)=
\mathbb E_{(c_t,A_t)\sim\mathcal D,\; z\sim p(z\mid A_t)}
\Big[\ell_{\mathrm{obj}}\big(\mathcal H_\theta(c_t,z),\,y(A_t,z)\big)\Big],
\]
and the usual action objectives are recovered by choosing \((z,y,\ell_{\mathrm{obj}})\):

\begin{itemize}[leftmargin=*,nosep]
    \item \textbf{Direct regression.} \(z=\varnothing\) and \(y=A_t\), with \(\ell_{\mathrm{obj}}\) a squared or Huber error; the chunk is emitted in one pass.
    \item \textbf{Iterative denoising.} \(z=(\tau,x_\tau)\) pairs a noise level with a corrupted chunk and \(y\) is the corresponding velocity or noise, with \(\ell_{\mathrm{obj}}\) a squared error. Flow matching and diffusion differ only in the corruption schedule and the target parameterization.
    \item \textbf{Autoregressive decoding.} \(z\) is a prefix of already-emitted tokens of a discretized chunk, \(y\) is the next token, and \(\ell_{\mathrm{obj}}\) is a cross-entropy.
\end{itemize}

All results reported here use a conditional flow-matching instantiation~\citep{lipman2022flow}, written out in \Cref{app:head}; we therefore make no empirical claim about the remaining objectives.
Actions and proprioception are normalized per dimension with benchmark training statistics and transformed back to native controls before execution.
Benchmark-specific optimization, hyperparameter, and control-interface details are provided in \Cref{tab:hparams,tab:interfaces}.

\section{Evaluation}
\label{sec:evaluation}

\subsection{Experimental Setup}
\label{sec:setup}
\paragraph{Benchmarks.}
\textbf{LIBERO} contains four ten-task suites and reports suite and macro success~\citep{liu2023libero}.
\textbf{LIBERO-PRO} augments these suites with controlled changes to object attributes, initial positions, semantically equivalent instructions, and task goals~\citep{zhou2025liberopro}.
\textbf{CALVIN} ABC$\rightarrow$D evaluates 1,000 chains of five subtasks in held-out scene D and reports SR@1--5 and average completed chain length~\citep{mees2022calvin}.
\textbf{SimplerEnv} evaluates \textbf{BridgeData V2} training on four \textbf{WidowX} tasks and Fractal training on Google Robot variant aggregations~\citep{li2024evaluating,walke2023bridgedata,brohan2022rt}.
\textbf{RoboTwin 2.0} evaluates bimanual control under the standard Easy/Hard (clean/randomized) settings~\citep{chen2025robotwin}.
\textbf{BEHAVIOR-1K} covers long-horizon household activities~\citep{BEHAVIOR1K}; we report the challenge's official Q-Score (computed internally as the per-task mean over rollouts, \texttt{q\_mean}).
We follow the standard protocol for each of the seven benchmark evaluations, which span six task domains and their corresponding control interfaces.
Complete protocol, optimization, hyperparameter, and control-interface details are provided in \Cref{tab:protocols,tab:hparams,tab:interfaces}; baseline values follow the corresponding cited sources.
In the results below we lead with BEHAVIOR-1K, the most demanding setting in this suite, and then follow the benchmark order above.

\begin{wrapfigure}[34]{R}{0.4\linewidth}
\centering
\includegraphics[width=0.94\linewidth]{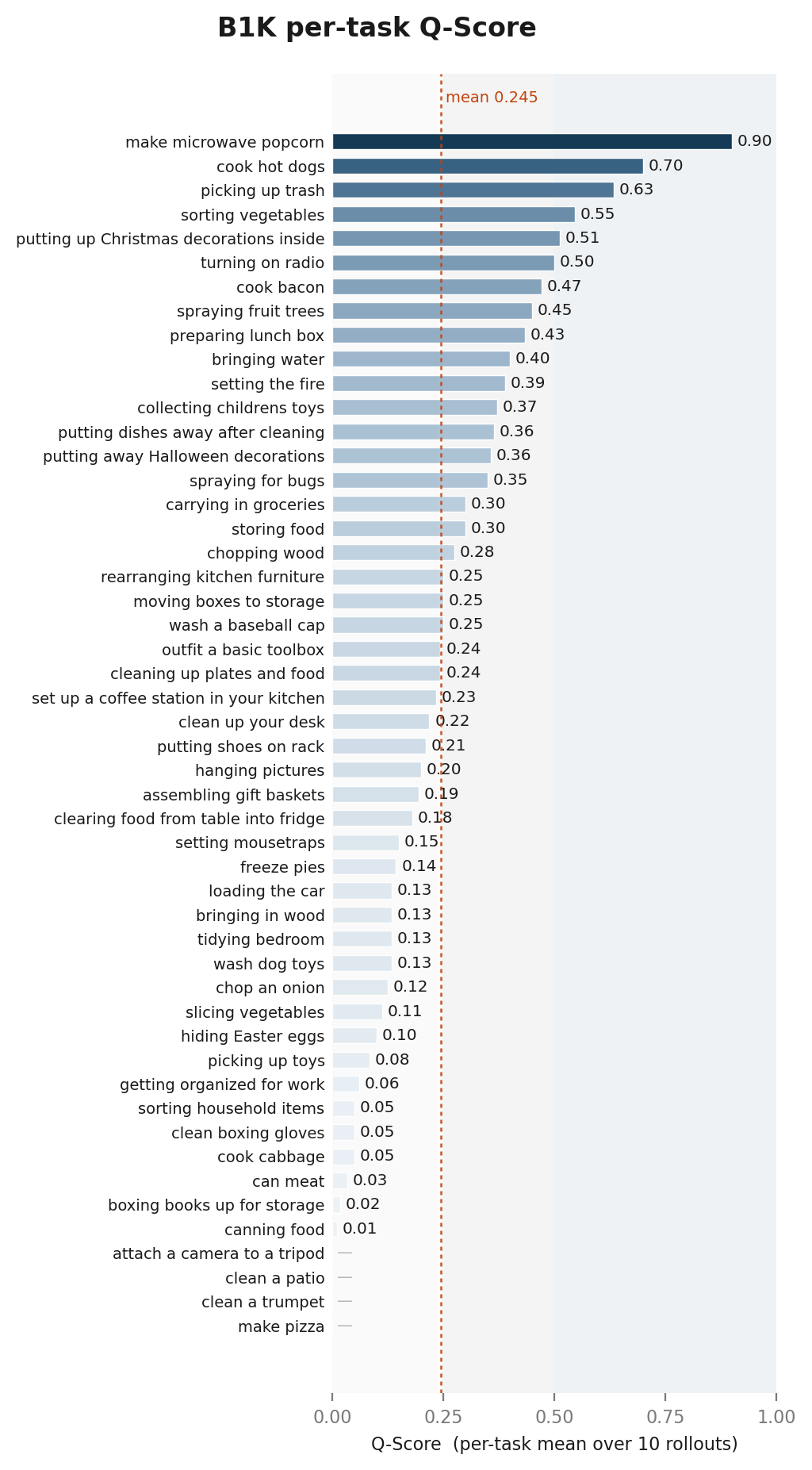}
\caption{\textbf{BEHAVIOR-1K \method{} per-task results.}
Per-task Q-Score over 10 rollouts for all evaluated household behaviors (dashed line: mean \(0.245\)).
Q-Score is the challenge's official partial-credit metric; internally we compute it as
\texttt{q\_mean}, the per-task mean over rollouts. }
\label{fig:b1k}
\end{wrapfigure}

\subsection{Headline Results}

\paragraph{BEHAVIOR-1K: Long-Horizon Household Behavior.}
BEHAVIOR-1K is the most demanding setting in our suite: the 2025 BEHAVIOR Challenge scores 50 full-length household activities in OmniGibson and supplies 10{,}000 teleoperated expert demonstrations for training~\citep{BEHAVIOR1K}.
Those demonstrations average \(397\) seconds and carry \(27.06\) annotated skill segments per trajectory drawn from 31 unique skills.
\Cref{fig:b1k} reports per-task Q-Score for all 50 tasks. \rev{\method{} achieves \(0.2450\) on Public Validation, which places second against the archived Standard-track scores} (\Cref{tab:behavior_leaderboard_compare}).
Q-Score is the challenge's official partial-credit metric over satisfied BDDL goal predicates, computed internally as \texttt{q\_mean}, the per-task mean over rollouts.

Its expert demonstrations average \(6.6\) minutes, while the longest activities extend to \(14.4\) minutes and approximately \(25{,}900\) visual frames.
Despite these horizons, \method{} obtains non-zero Q-Score on eight of the nine activities whose expert demonstrations exceed nine minutes.
Two long multi-stage activities outperform the overall 50-task mean: \emph{storing food} reaches \(0.30\) over an 11.1-minute demonstration horizon (approximately \(20{,}000\) frames), and \emph{collecting children's toys} reaches \(0.37\) over 10.6 minutes (approximately \(19{,}000\) frames).
The former requires a multi-room sequence---retrieving groceries from a car trunk in the garage, carrying them to the kitchen, placing items in the refrigerator, and closing both the trunk and refrigerator.
These outcomes show that \method{} can sustain meaningful progress through long, multi-stage household activities using onboard observations, without a pretrained language model, explicit subgoal decomposition, or test-time search.
\Cref{fig:b1k_rollouts} shows goal-satisfying episodes on eight further activities, including \emph{cook bacon} and \emph{setting the fire}.

\begin{table}[t]
  \centering
  % \vspace{-0.2 in}
  \caption{\textbf{BEHAVIOR-1K leaderboard comparison.}
  \rev{Comparison of our Public Validation Q-Score with the archived 2025 challenge leaderboard scores.} \method{} is highlighted in bold. All rows are Standard track except
  Embodied Intelligence ($^\dagger$Privileged).}
  \label{tab:behavior_leaderboard_compare}
  \scriptsize
  \resizebox{\ifdim\width>\linewidth\linewidth\else\width\fi}{!}{%
  \begin{tabular}{clc}
    \toprule
    Rank & Team & Q-Score (Public Val.) \\
    \midrule
    1 & Robot Learning Collective~\citep{b1k2025:rlc} & 0.2605 \\
    \rev{\textbf{(2)}} & \textbf{\method{} (Ours)} & \textbf{0.2450} \\
    3 & SimpleAI Robot~\citep{b1k2025:simpleai} & 0.1943 \\
    4 & Comet~\citep{b1k2025:comet} & 0.1830 \\
    5 & The North Star~\citep{b1k2025:northstar} & 0.1702 \\
    6 & Embodied Intelligence$^\dagger$~\citep{b1k2025:ei} & 0.1110 \\
    \bottomrule
  \end{tabular}}
\end{table}

\paragraph{LIBERO and CALVIN.}
On LIBERO, \method{} achieves \(96.4\%\) average success across the four task suites, including \(91.4\%\) on LIBERO-Long, placing the LM-free policy among the strongest recent VLA systems (\Cref{tab:libero}). On CALVIN ABC$\rightarrow$D, it reaches an average completed chain length of \(4.24\) out of five in the held-out scene, demonstrating sustained multi-step execution under instruction-conditioned control (\Cref{tab:calvin}).
Together, these results show that pretrained language-side parameters are not required for high success across either diverse manipulation suites or compositional multi-step sequences.

\begin{table}[t]
\centering
\begin{minipage}[t]{0.50\linewidth}
\centering
\captionof{table}{\textbf{LIBERO comparison.} Success (\%).
``B'' is the scale of the \emph{pretrained} backbone in billions.}
\label{tab:libero}
\scriptsize
\setlength{\tabcolsep}{2pt}
\resizebox{0.95\linewidth}{!}{%
\begin{tabular}{l|c|cccc|c}
\toprule
\textbf{Model} & \textbf{B} & \textbf{Spatial} & \textbf{Object} & \textbf{Goal} & \textbf{Long} & \textbf{Avg} \\
\midrule
\multicolumn{7}{l}{\textit{\textbf{Large Models ($\ge$ 4B)}}} \\
\midrule
UniVLA \citep{UniVLA} & 9 & 96.5 & 96.8 & 95.6 & 92.0 & 95.2 \\
FlowVLA \citep{FlowVLA} & 8.5 & 93.2 & 95.0 & 91.6 & 72.6 & 88.1 \\
UnifiedVLA \citep{UnifiedVLA} & 8.5 & 95.4 & 98.8 & 93.6 & 94.0 & 95.5 \\
OpenVLA \citep{kim2024openvla} & 7 & 84.7 & 88.4 & 79.2 & 53.7 & 76.5 \\
OpenVLA-OFT \citep{OpenVLA_OFT} & 7 & 97.6 & 98.4 & 97.9 & 94.5 & 97.1 \\
DD-VLA \citep{DD_VLA} & 7 & 97.2 & 98.6 & 97.4 & 92.0 & 96.3 \\
MemoryVLA \citep{MemoryVLA} & 7 & 98.4 & 98.4 & 96.4 & 93.4 & 96.7 \\
PD-VLA \citep{PD_VLA} & 7 & 95.5 & 96.7 & 94.9 & 91.7 & 94.7 \\
MolmoAct \citep{MolmoAct} & 7 & 87.0 & 95.4 & 87.6 & 77.2 & 86.8 \\
ThinkAct \citep{ThinkAct} & 7 & 88.3 & 91.4 & 87.1 & 70.9 & 84.4 \\
CoT-VLA \citep{CoT_VLA} & 7 & 87.5 & 91.6 & 87.6 & 69.0 & 83.9 \\
WorldVLA \citep{WorldVLA} & 7 & 87.6 & 96.2 & 83.4 & 60.0 & 81.8 \\
TraceVLA \citep{TraceVLA} & 7 & 84.6 & 85.2 & 75.1 & 54.1 & 74.8 \\
FPC-VLA \citep{FPC_VLA} & 7 & 86.2 & 87.0 & 92.0 & 82.2 & 86.9 \\
4D-VLA \citep{p4D_VLA} & 4 & 88.9 & 95.2 & 90.9 & 79.1 & 88.5 \\
SpatialVLA \citep{SpatialVLA} & 4 & 88.2 & 89.9 & 78.6 & 55.5 & 78.1 \\
\midrule
\multicolumn{7}{l}{\textit{\textbf{Small Models ($<$ 4B)}}} \\
\midrule
$\pi_0$ \citep{pi0} & 3 & 96.8 & 98.8 & 95.8 & 85.2 & 94.2 \\
$\pi_0$-FAST \citep{FAST} & 3 & 96.4 & 96.8 & 88.6 & 60.2 & 85.5 \\
$\pi_{0.5}$ \citep{pi05} & 3 & 98.8 & 98.2 & 98.0 & 92.4 & 96.9 \\
NORA \citep{NORA} & 3 & 92.2 & 95.4 & 89.4 & 74.6 & 87.9 \\
SmolVLA \citep{SmolVLA} & 2.2 & 93.0 & 94.0 & 91.0 & 77.0 & 88.8 \\
VLANeXt \citep{wu2026vlanext} & 2 & 99.0 & 99.2 & 96.6 & 94.6 & 97.4 \\
GR00T-N1 \citep{bjorck2025gr00t} & 2 & 94.4 & 97.6 & 93.0 & 90.6 & 93.9 \\
GraspVLA \citep{GraspVLA} & 1.8 & - & 94.1 & 91.2 & 82.0 & 89.1 \\
FLOWER \citep{FLOWER} & 1 & 97.5 & 99.1 & 96.1 & 94.9 & 96.9 \\
\midrule
\multicolumn{7}{l}{\textit{\textbf{Tiny Models ($<$ 1B)}}} \\
\midrule
X-VLA \citep{x_VLA} & 0.9 & 98.2 & 98.6 & 97.8 & 97.6 & 98.1 \\
VLA-Adapter \citep{VLA_Adapter} & 0.5 & 97.8 & 99.2 & 97.2 & 95.0 & 97.3 \\
VLA-OS \citep{VLA_OS} & 0.5 & 87.0 & 96.5 & 92.7 & 66.0 & 85.6 \\
UniAct \citep{UniAct} & 0.5 & 77.0 & 87.0 & 77.0 & 70.0 & 77.8 \\
SimVLA \citep{luo2026simvla} & 0.5 & 99.6 & 99.8 & 98.6 & 96.4 & 98.6 \\
\midrule
\rowcolor{resultrow}
\method{} & 0.3 & 98.6 & 99.0 & 96.4 & 91.4 & 96.4 \\
\bottomrule
\end{tabular}}
\end{minipage}\hfill
\begin{minipage}[t]{0.48\linewidth}
\centering
\captionof{table}{\textbf{CALVIN ABC$\rightarrow$D comparison.} Success (\%) and average chain length.}
\label{tab:calvin}
\scriptsize
\setlength{\tabcolsep}{3pt}
\resizebox{\linewidth}{!}{%
\begin{tabular}{l|ccccc|c}
\toprule
\textbf{Model} & \textbf{SR@1} & \textbf{SR@2} & \textbf{SR@3} & \textbf{SR@4} & \textbf{SR@5} & \textbf{Avg.\ len} \\
\midrule
\multicolumn{7}{l}{\textit{\textbf{Large ($\ge$ 4B)}}} \\
\midrule
UniVLA \citep{UniVLA} & 95.5 & 85.8 & 75.4 & 66.9 & 56.5 & 3.80 \\
UnifiedVLA \citep{UnifiedVLA} & 98.9 & 94.8 & 89.0 & 82.8 & 75.1 & 4.41 \\
OpenVLA \citep{kim2024openvla} & 91.3 & 77.8 & 62.0 & 52.1 & 43.5 & 3.27 \\
OpenVLA-OFT \citep{OpenVLA_OFT} & 96.3 & 89.1 & 82.4 & 75.8 & 66.5 & 4.10 \\
HiF-VLA \citep{HiF-VLA} & 98.5 & 94.1 & 88.1 & 81.4 & 73.1 & 4.35 \\
\midrule
\multicolumn{7}{l}{\textit{\textbf{Tiny ($<$ 1B)}}} \\
\midrule
Seer \citep{Seer} & 96.3 & 91.6 & 86.1 & 80.3 & 74.0 & 4.28 \\
VLA-Adapter \citep{VLA_Adapter} & 99.1 & 94.6 & 88.8 & 82.8 & 76.5 & 4.42 \\
X-VLA \citep{x_VLA} & 97.1 & 92.6 & 88.5 & 84.4 & 78.8 & 4.43 \\
SimVLA \citep{luo2026simvla} & 99.4 & 96.4 & 89.8 & 83.7 & 77.2 & 4.47 \\
\midrule
\rowcolor{resultrow}
\method{} & 99.4 & 93.0 & 85.8 & 78.6 & 67.6 & 4.24 \\
\bottomrule
\end{tabular}}

\vspace{1.2em}

\captionof{table}{\textbf{RoboTwin 2.0 comparison.} Success (\%) under the standard Easy/Hard settings (\texttt{eval\_clean}/\texttt{eval\_randomized}). }
% $^\ddagger$ is the unweighted mean of the two evaluation settings. }
\label{tab:robotwin}
\scriptsize
\setlength{\tabcolsep}{3pt}
\resizebox{0.98\linewidth}{!}{%
\begin{tabular}{l|cc|c}
\toprule
\textbf{Model} & \textbf{Easy} & \textbf{Hard} & \textbf{Avg} \\
\midrule
$\pi_0$ \citep{pi0} & 65.9 & 58.4 & 62.2 \\
$\pi_{0.5}$ \citep{pi05} & 82.7 & 76.8 & 79.8 \\
LingBot-VLA \citep{LingBotVLA} & 86.5 & 85.3 & 85.9 \\
Motus \citep{Motus} & 88.7 & 87.0 & 87.8 \\
StarVLA \citep{StarVLA} & 88.7 & 87.8 & 88.3 \\
Fast-WAM \citep{FastWAM} & 91.9 & 91.8 & 91.8 \\
LingBot-VA \citep{LingBotVA} & 92.9 & 91.5 & 92.2 \\
G0.5 \citep{GalaxeaG05} & 93.7 & 92.8 & 93.3 \\
\midrule
\rowcolor{resultrow}
\method{} & 94.0 & 87.2 & 90.6 \\
\bottomrule
\end{tabular}}
\end{minipage}
\end{table}

\paragraph{LIBERO-PRO.}
\rev{\Cref{tab:libero_pro} reports success rates by perturbation type.}
Most notably, it achieves \(96.5\%\) under semantic instruction perturbations, \rev{near its standard LIBERO average of \(96.4\%\)} despite loading no pretrained language parameters.
\rev{Object-attribute success averages \(80.0\%\) (\(54.0\%\) on Long), compared with \(96.0\%\) for \(\pi_{0.5}\) and \(93.0\%\) for OpenVLA. For position perturbations, our \(21.5\%\) is close to the \(20.75\%\) of \(\pi_{0.5}\).}
Recomposed task goals \rev{yield low success across methods} (\(\pi_{0.5}\) \(0.8\%\), SimVLA \(6.0\%\), and \method{} \(6.5\%\)).

\begin{table}[t]
\centering
\vspace{-0.05 in}
\begin{minipage}[t]{0.48\linewidth}
\centering
\captionof{table}{\textbf{WidowX comparison.} Success (\%).}
\label{tab:widowx}
\scriptsize
\setlength{\tabcolsep}{2pt}
\resizebox{\linewidth}{!}{%
\begin{tabular}{l|cccc|c}
\toprule
\textbf{Method} & \textbf{Spoon} & \textbf{Carrot} & \textbf{Stack} & \textbf{Eggplant} & \textbf{Avg} \\
\midrule
\multicolumn{6}{l}{\textit{\textbf{Large ($\ge$ 4B)}}} \\ \midrule
RT-1-X \citep{o2024open} & 0.0 & 4.2 & 0.0 & 0.0 & 1.1 \\
Octo-Base \citep{Octo} & 12.5 & 8.3 & 0.0 & 43.1 & 16.0 \\
OpenVLA \citep{kim2024openvla} & 0.0 & 0.0 & 0.0 & 4.1 & 1.0 \\
OpenVLA-OFT \citep{OpenVLA_OFT} & 12.5 & 4.2 & 8.3 & 37.5 & 15.6 \\
DD-VLA \citep{DD_VLA} & 29.2 & {29.2} & {20.8} & {70.8} & {37.5} \\
RoboVLM \citep{li2024towards} & 29.2 & 25.0 & 12.5 & 58.3 & 31.3 \\
SpatialVLA \citep{SpatialVLA} & 16.7 & 25.0 & 29.2 & 100.0 & 42.7 \\
MemoryVLA \citep{MemoryVLA} & 75.0 & 75.0 & 37.5 & 100.0 & 71.9 \\
ThinkAct \citep{ThinkAct} & 58.3 & 37.5 & 8.7 & 70.8 & 43.8 \\
FPC-VLA \citep{FPC_VLA} & 79.2 & 58.3 & 45.8 & 75.0 & 64.6 \\
\midrule
\multicolumn{6}{l}{\textit{\textbf{Small ($<$ 4B)}}} \\ \midrule
Octo-Small \citep{Octo} & {47.2} & 9.7 & 4.2 & 56.9 & 29.5 \\
$\pi_0$ \citep{pi0} & 29.1 & 0.0 & {16.7} & 62.5 & 27.1 \\
$\pi_0$-FAST \citep{FAST} & 29.1 & 21.9 & 10.8 & {66.6} & 32.1 \\
GR00T-N1 \citep{bjorck2025gr00t} & 62.5 & 45.8 & {16.7} & 20.8 & {36.5} \\
FLOWER \citep{FLOWER} & 71.0 & 13.0 & 8.0 & 88.0 & 45.0 \\
\midrule
\multicolumn{6}{l}{\textit{\textbf{Tiny ($<$ 1B)}}} \\ \midrule
X-VLA \citep{x_VLA} & 100.0 & 91.7 & 95.8 & 95.8 & 95.8 \\
SimVLA \citep{luo2026simvla} & 100.0 & 100.0 & 91.7 & 91.7 & 95.8 \\
\rowcolor{resultrow}
\method{} & 100.0 & 100.0 & 79.2 & 87.5 & 91.7 \\
\bottomrule
\end{tabular}}
\end{minipage}\hfill
\begin{minipage}[t]{0.50\linewidth}
\centering
\captionof{table}{\textbf{Google Robot comparison.} Success (\%). }
\label{tab:google}
\scriptsize
\setlength{\tabcolsep}{2pt}
\resizebox{0.9\linewidth}{!}{%
\begin{tabular}{l|ccc|c}
\toprule
\textbf{Model} & \textbf{Pick} & \textbf{Move} & \textbf{Open} & \textbf{Avg} \\
\midrule
\multicolumn{5}{l}{\textit{\textbf{Large ($\ge$ 4B)}}} \\
\midrule
RT-1-X \citep{o2024open} & 49.0 & 32.3 & 29.4 & 36.9 \\
RT-2-X \citep{o2024open} & 82.3 & 79.2 & 35.3 & 65.6 \\
Octo-Base \citep{Octo} & 0.6 & 3.1 & 1.1 & 1.6 \\
OpenVLA \citep{kim2024openvla} & 54.5 & 47.7 & 17.7 & 40.0 \\
OpenVLA-OFT \citep{OpenVLA_OFT} & 65.3 & 59.0 & 12.2 & 45.5 \\
RoboVLM \citep{li2024towards} & 75.6 & 60.0 & 10.6 & 48.7 \\
TraceVLA \citep{TraceVLA} & 60.0 & 56.4 & 31.0 & 49.1 \\
DD-VLA \citep{DD_VLA} & 82.5 & 64.6 & 23.6 & 56.9 \\
SpatialVLA \citep{SpatialVLA} & 88.0 & 72.7 & 41.8 & 67.5 \\
ThinkAct \citep{ThinkAct} & 84.0 & 63.8 & 47.6 & 65.1 \\
\midrule
\multicolumn{5}{l}{\textit{\textbf{Small ($<$ 4B)}}} \\
\midrule
$\pi_0$ \citep{pi0} & 75.2 & 63.7 & 25.6 & 54.8 \\
$\pi_0$-FAST \citep{FAST} & 77.6 & 68.2 & 31.3 & 59.0 \\
GR00T-N1 \citep{bjorck2025gr00t} & 78.8 & 62.5 & 13.2 & 51.5 \\
\midrule
\multicolumn{5}{l}{\textit{\textbf{Tiny ($<$ 1B)}}} \\
\midrule
X-VLA \citep{x_VLA} & 85.5 & 79.8 & 61.9 & 75.7 \\
SimVLA \citep{luo2026simvla} & 82.3 & 81.0 & 67.7 & 77.0 \\
\rowcolor{resultrow}
\method{} & 89.7 & 76.5 & 79.9 & 82.0 \\
\bottomrule
\end{tabular}}
\end{minipage}
\end{table}

\paragraph{RoboTwin 2.0.}
\method{} achieves \(94.0\%\) success in the Easy setting and \(87.2\%\) under randomized Hard evaluation, yielding a \(90.6\%\) average across the two settings (\Cref{tab:robotwin}).
The Easy result is the highest in the comparison, while the Hard result remains competitive with leading vision-language-action and video-action policies.
This performance extends the LM-free construction from single-arm manipulation to coordinated bimanual control under visual and object-level variation.

\paragraph{WidowX and Google Robot.}
\method{} achieves \(91.7\%\) average success on WidowX, including perfect success on Spoon on Towel and Carrot on Plate, and ranks third among the compared methods (\Cref{tab:widowx}).
\rev{It reaches} \(79.2\%\) on Stack Cube and \(87.5\%\) on Eggplant in Basket.
On Google Robot, it reaches an \(82.0\%\) variant-aggregation average, the strongest result among all methods in \Cref{tab:google}.
These evaluations extend the LM-free result across distinct robot embodiments, visual domains, and controller semantics in SimplerEnv.

\paragraph{Comparison against the language-pretrained construction.}
Among the tabulated systems, SimVLA is \rev{a system-level reference that retains} the pretrained language side of a vision--language backbone~\citep{luo2026simvla}.
\rev{The two systems use the same source observations and search space for shared hyperparameters, with differences in backbones, input resizing, and selected configurations (\Cref{app:simvla_removal}).}
Across the shared evaluations the margin is small and does not favor one side uniformly: \method{} trails by \rev{\(2.2\)} points in LIBERO average success, by \(0.23\) in CALVIN average chain length, while exceeding the language-pretrained configuration by \(5.0\) points on Google Robot.
Under LIBERO-PRO the two remain within \(2.3\) points on semantic instruction perturbations (\(96.5\%\) versus \(98.8\%\)), whereas \method{} is substantially stronger under initial-position perturbation (\(21.5\%\) versus \(8.3\%\)).
These results are obtained with a \(0.3\)B vision backbone in place of a \(0.5\)B vision--language backbone and with no loaded language parameters.
\rev{These comparisons establish competitive sufficiency.}

\subsection{Ablations}
\label{sec:ablation}
This section identifies which components of the construction are responsible for the reported performance (\Cref{tab:ablations}).
The no-instruction variant removes the instruction pathway.
We additionally compare the default random, action-trained instruction embeddings with pretrained language-side alternatives (token embedding table and full text encoder) derived from BERT~\citep{devlin2019bert} and SigLIP2~\citep{tschannen2025siglip2}.
On the vision side, the DINOv3 variant replaces SigLIP2 with a vision encoder pretrained without text supervision~\citep{simeoni2025dinov3}, and the no-vision-pretraining variant trains the same SigLIP2 encoder architecture from random initialization.
\Cref{tab:ablations} reports all component scores and their mean in percent.

\begin{table*}[t]
\centering
% \vspace{-0.2 in}
\begin{minipage}[t]{0.49\textwidth}
\vspace{0pt}
\centering
\captionof{table}{\textbf{Robustness evaluation on LIBERO-PRO.}
Success rate (\%) on the original tasks (Ori) and under object-attribute (Obj), initial-position (Pos), semantic-instruction (Sem), and task-goal (Task) perturbations~\citep{zhou2025liberopro}.}
% \vspace{0.1 in}
\label{tab:libero_pro}
\scriptsize
\setlength{\tabcolsep}{2.5pt}
\resizebox{\linewidth}{!}{%
\begin{tabular}{l|l|ccccc}
\toprule
\textbf{Suite} & \textbf{Method} & \textbf{Ori} & \textbf{Obj} & \textbf{Pos} & \textbf{Sem} & \textbf{Task} \\
\midrule
& OpenVLA \citep{kim2024openvla} & 98.0 & 97.0 & 0.0 & 97.0 & 0.0 \\
& $\pi_{0}$ \citep{pi0} & 96.8 & 95.0 & 0.0 & 97.0 & 0.0 \\
& $\pi_{0.5}$ \citep{pi05} & 98.0 & 97.0 & 20.0 & 97.0 & 1.0 \\
& SimVLA \citep{luo2026simvla} & 99.0 & 98.0 & 29.0 & 98.0 & 0.0 \\
\rowcolor{resultrow} \multirow{-5}{*}{Spatial} & \method{} & 98.4 & 98.0 & 40.0 & 96.0 & 0.0 \\
\midrule
& OpenVLA \citep{kim2024openvla} & 99.0 & 98.0 & 0.0 & 98.0 & 0.0 \\
& $\pi_{0}$ \citep{pi0} & 98.8 & 94.0 & 0.0 & 90.0 & 0.0 \\
& $\pi_{0.5}$ \citep{pi05} & 98.0 & 98.0 & 17.0 & 96.0 & 1.0 \\
& SimVLA \citep{luo2026simvla} & 100.0 & 85.0 & 1.0 & 100.0 & 4.0 \\
\rowcolor{resultrow} \multirow{-5}{*}{Object} & \method{} & 99.6 & 90.0 & 28.0 & 100.0 & 10.0 \\
\midrule
& OpenVLA \citep{kim2024openvla} & 98.0 & 96.0 & 0.0 & 98.0 & 0.0 \\
& $\pi_{0}$ \citep{pi0} & 95.8 & 94.0 & 0.0 & 93.0 & 0.0 \\
& $\pi_{0.5}$ \citep{pi05} & 97.0 & 97.0 & 38.0 & 97.0 & 0.0 \\
& SimVLA \citep{luo2026simvla} & 99.0 & 82.0 & 0.0 & 99.0 & 10.0 \\
\rowcolor{resultrow} \multirow{-5}{*}{Goal} & \method{} & 97.4 & 78.0 & 4.0 & 96.0 & 10.0 \\
\midrule
& OpenVLA \citep{kim2024openvla} & 93.0 & 81.0 & 0.0 & 96.0 & 0.0 \\
& $\pi_{0}$ \citep{pi0} & 85.2 & 79.0 & 0.0 & 82.0 & 0.0 \\
& $\pi_{0.5}$ \citep{pi05} & 93.0 & 92.0 & 8.0 & 93.0 & 1.0 \\
& SimVLA \citep{luo2026simvla} & 96.0 & 61.0 & 3.0 & 98.0 & 10.0 \\
\rowcolor{resultrow} \multirow{-5}{*}{Long} & \method{} & 92.2 & 54.0 & 14.0 & 94.0 & 6.0 \\
\bottomrule
\end{tabular}}
\end{minipage}\hfill
\begin{minipage}[t]{0.49\textwidth}
\vspace{0pt}
\centering
\captionof{table}{\textbf{Ablations on LIBERO and WidowX (\%).}
The reported \method{} result is the default configuration; each ablation changes one component while retaining the remaining architecture and training recipe.}
\label{tab:ablations}
\scriptsize
\setlength{\tabcolsep}{3pt}
\textbf{(a) LIBERO}\\
\resizebox{\linewidth}{!}{%
\begin{tabular}{l|l|ccccc}
\toprule
\textbf{Knob} & \textbf{Value} & \textbf{Spatial} & \textbf{Object} & \textbf{Goal} & \textbf{Long} & \textbf{Avg.} \\
\midrule
\rowcolor{resultrow}
\textbf{\method{}} & \textbf{Default} & \textbf{98.6} & \textbf{99.0} & \textbf{96.4} & \textbf{91.4} & \textbf{96.4} \\
\midrule
Instruction conditioning & None & 75.4 & 93.4 & 6.4 & 15.2 & 47.6 \\
\midrule
\multirow{3}{*}{Text embedding} & BERT emb. table & 99.8 & 99.0 & 97.6 & 90.8 & 96.8 \\
  % & BERT text encoder &  &  &  &  & \\
 & SigLIP2 emb. table & 98.8 & 99.2 & 96.4 & 84.4 & 94.7 \\
   & SigLIP2 text encoder & 96.2 & 98.8 & 88.2 & 82.0 & 91.3 \\
\midrule
\multirow{2}{*}{Vision backbone} & DINOv3 & 95.6 & 95.0 & 89.4 & 80.2 & 90.1 \\
 & No vision pretraining & 69.2 & 85.0 & 66.8 & 16.2 & 59.3 \\
\bottomrule
\end{tabular}}

\vspace{0.65em}
\textbf{(b) WidowX}\\
\resizebox{\linewidth}{!}{%
\begin{tabular}{l|l|ccccc}
\toprule
\textbf{Knob} & \textbf{Value} & \textbf{Carrot} & \textbf{Eggplant} & \textbf{Spoon} & \textbf{Stack} & \textbf{Avg.} \\
\midrule
\rowcolor{resultrow}
\textbf{\method{}} & \textbf{Default} & \textbf{100.0} & \textbf{87.5} & \textbf{100.0} & \textbf{79.2} & \textbf{91.7} \\
\midrule
Instruction conditioning & None & 83.3 & 20.8 & 91.7 & 41.7 & 59.4 \\
\midrule
\multirow{3}{*}{Text embedding} & BERT emb. table & 100.0 & 91.7 & 95.8 & 75.0 & 90.6 \\
  % & BERT text encoder &  &  &  &  & \\
 & SigLIP2 emb. table & 95.8 & 50.0 & 100.0 & 83.3 & 82.3 \\
  
  & SigLIP2 text encoder & 95.8 & 41.7 & 100.0 & 83.3 & 80.2 \\
 \midrule
\multirow{2}{*}{Vision backbone} & DINOv3 & 91.7 & 0.0 & 87.5 & 8.3 & 46.9 \\
 & No vision pretraining & 16.7 & 4.2 & 25.0 & 0.0 & 11.5 \\
\bottomrule
\end{tabular}}
\end{minipage}
\end{table*}

\paragraph{Instruction conditioning provides the task signal.}
Removing instruction conditioning reduces LIBERO average success from \(96.4\%\) to \(47.6\%\) and WidowX average success from \(91.7\%\) to \(59.4\%\).
On LIBERO, the largest reductions occur on Goal and Long, whereas Object remains at \(93.4\%\), indicating that visual observations can identify some object-centric behaviors but are insufficient when the same scene admits distinct goals or extended task sequences.
The result directly separates the role of instruction \emph{conditioning} from language \emph{pretraining}: pretrained language parameters are unnecessary, but the action-trained instruction codes carry essential task information.

\paragraph{Pretrained text representations provide \rev{no consistent advantage.}}
We test two levels of language-side pretraining while holding the vision encoder, action Transformer, data, and optimization fixed.
The embedding-table variants replace the random lookup table with BERT's or SigLIP2's pretrained token embeddings, which remain trainable under the action objective; the text-encoder variant instead feeds the contextual token outputs of the full pretrained SigLIP2 text tower to the action Transformer.
The BERT table reaches \(96.8\%\) on LIBERO and \(90.6\%\) on WidowX; the SigLIP2 table reaches \(94.7\%\) and \(82.3\%\), and the full SigLIP2 text encoder \(91.3\%\) and \(80.2\%\).

\paragraph{Visual pretraining matters.}
Replacing SigLIP2 with DINOv3 retains \(90.1\%\) average success on LIBERO, although the WidowX average falls to \(46.9\%\), suggesting that WidowX may benefit more strongly from language-aligned visual pretraining.
Randomly initializing and training the same SigLIP2 vision architecture from scratch reduces performance further to \(59.3\%\) on LIBERO and \(11.5\%\) on WidowX.
Thus, strong visual pretraining is important.

\paragraph{\rev{Execution under an unseen instruction string.}}
The WidowX instruction ``put eggplant into yellow basket'' \rev{does not appear as a complete string in the training instructions}, although all of its tokens are individually covered by the training vocabulary.
\method{} reaches \(87.5\%\) on this task, compared with \(20.8\%\) for the no-instruction variant.
\rev{This shows successful execution under an unseen instruction string composed of training-seen tokens. See \Cref{app:instruction_diagnostics,app:whole_instruction} for instruction diagnostics.}
\section{Conclusion}
\label{sec:conclusion}

\method{} keeps instruction conditioning while removing all pretrained language-side parameters.
Across seven evaluations spanning six manipulation domains, it remains competitive with language-pretrained systems and \rev{achieves a Q-Score of \(0.2450\) on BEHAVIOR-1K Public Validation.}
The ablations show that instructions remain essential, pretrained text representations provide no consistent advantage, and visual pretraining still matters.
Thus, these task-bounded control benchmarks do not require pretrained language-side parameters.

\paragraph{Language remains the brain.}
Language remains central to open-ended instruction understanding, task decomposition, planning, reasoning, and interaction.
We therefore advocate a modular brain--cerebellum organization: a full VLM handles high-level semantics and reasoning, while \method{} executes bounded skills as \rev{an} instruction-conditioned controller.

\paragraph{Limitations.}
Our two 8$\times$H100 nodes did not permit \rev{exhaustive} per-domain hyperparameter search; BEHAVIOR-1K in particular uses a single heuristically chosen configuration, so further tuning may improve its Q-Score.
We evaluate only the instruction-conditioned controller; integrating it with a high-level VLM planner remains future work.

% \paragraph{Broader impact.}
% Identifying when language pretraining is unnecessary can reduce model complexity and sharpen VLA evaluation.
% Benchmark performance nevertheless does not imply safe deployment, which still requires physical safeguards and human oversight.

% \subsection*{Reproducibility Statement}
% \Cref{sec:setup} summarizes the benchmark suite and evaluation setup.
% The appendix provides evaluation protocols, benchmark-specific hyperparameters, control interfaces, and flow-matching implementation details (\Cref{tab:protocols}; \Cref{tab:hparams}; \Cref{tab:interfaces}; \Cref{app:head}).
% Code is available at \url{https://github.com/LUOyk1999/SimVLA2}.

% \subsection*{Ethics Statement}
% The reported benchmarks are simulated or benchmark-mediated robot-control evaluations and contain no human-subject data.
% Benchmark results do not establish safe deployment; physical robot systems require collision handling, uncertainty monitoring, workspace limits, hardware interlocks, and human oversight.

\subsection*{AI Use Statement}
Generative AI tools assisted manuscript restructuring and language editing.
They were not used to generate experimental measurements.
The authors independently reviewed the experimental results, implementation, claims, and citations and remain responsible for the final paper and artifacts.

% \clearpage
\bibliographystyle{iclr2027_conference}
\bibliography{simva_references}

\newpage
\appendix
\section{Implementation Details}
\label{app:implementation}

\subsection{Evaluation Protocols}
\Cref{tab:protocols} summarizes the evaluator setting and metric used for each benchmark evaluation.

\begin{table}[h]
\centering
\caption{\textbf{Benchmark-specific evaluation protocols.}}
\label{tab:protocols}
\scriptsize
\setlength{\tabcolsep}{3pt}
\resizebox{\linewidth}{!}{%
\begin{tabular}{l|l|l}
\toprule
\textbf{Evaluation} & \textbf{Protocol} & \textbf{Metric} \\
\midrule
LIBERO & Four ten-task suites; predict \(H{=}10\), execute \(h{=}5\) & Suite and macro success (\%) \\
LIBERO-PRO & Original plus object, position, semantic, and task perturbations & Per-suite and perturbation success (\%) \\
CALVIN ABC\(\rightarrow\)D & 1{,}000 five-subtask chains in held-out scene D & SR@1--5, mean chain length \\
WidowX & SimplerEnv-Bridge, four manipulation tasks & Per-task success (\%), average \\
Google Robot & SimplerEnv-Fractal variant aggregation & Variant-aggregation success (\%) \\
RoboTwin 2.0 & Bimanual Easy/Hard evaluation & Success (\%) \\
BEHAVIOR-1K \rev{Public Validation} & 50 household behaviors, 10 rollouts each & Q-Score \\
\bottomrule
\end{tabular}}
\end{table}

\subsection{Domain Interfaces and Action Spaces}
\Cref{tab:interfaces} makes explicit what is and is not shared across domains.
All runs use the high-level \(\{\)images, view mask, instruction codes, proprioception, action\(\}\) interface, with benchmark-specific dataset statistics, control adapters, and evaluators.

\begin{table}[h]
\caption{\textbf{Benchmark-specific control interfaces.}
\(H\) is predicted chunk length, \(h\) is maximum execution before replanning, \(r\) is execution stride, and \(K\) is the Euler-step count.}
\label{tab:interfaces}
\centering
\scriptsize
\setlength{\tabcolsep}{2.6pt}
\resizebox{\linewidth}{!}{%
\begin{tabular}{lccccl}
\toprule
Domain & Views & State & Native/model action & \(H/h/r/K\) & Control semantics \\
\midrule
LIBERO/LIBERO-PRO & 2 & 8D & 7D/7D & 10/5/1/10 & Delta EE pose/gripper \\
CALVIN & 2 & 20D & 10D/20D & 30/30/1/20 & 10D active arm + 10D zero dummy; execute first 10D \\
WidowX & 1 & 7D & 7D/7D & 30/30/1/30 & Absolute EE position/Euler target \\
Google Robot & 1 & zero 7D & 7D/7D & 30/20/1/30 & Accumulated translation; absolute orientation \\
RoboTwin 2.0 & 3 & 14D & 14D/14D & 125/50/1/10 & Absolute dual-arm joint targets \\
BEHAVIOR-1K & 3 & 23D & 23D/23D & 30/30/1/10 & Relative targets for trunk/arms; absolute base velocity and grippers \\
\bottomrule
\end{tabular}}
\end{table}

\subsection{Training Hyperparameters}
\label{app:training}
\Cref{tab:hparams} provides the benchmark-specific schedules used for the reported manipulation results.
Training is iteration-based: because trajectories have different frame-level lengths across datasets, a nominal epoch is not directly comparable across domains, so we report the total number of training iterations.

\begin{table}[h]
\caption{\textbf{Benchmark-specific training hyperparameters.}
$^\S$BEHAVIOR-1K is trained in two stages: a 50-task multi-task pretraining run, followed by a fine-tuning run at the lower learning rate.}
\label{tab:hparams}
\centering
\scriptsize
\setlength{\tabcolsep}{2.5pt}
\resizebox{0.9\linewidth}{!}{%
\begin{tabular}{lcccccc}
\toprule
Configuration & LIBERO/LIBERO-PRO & CALVIN & WidowX & Google & RoboTwin & BEHAVIOR-1K \\
\midrule
Image size / valid views & 256 / 2 & 256 / 2 & 256 / 1 & 256 / 1 & 256 / 3 & 256 / 3 \\
Action head \(d_{\mathrm{model}}\)/layers/heads & 1024/24/16 & 1024/24/16 & 1024/24/16 & 1024/24/16 & 1024/24/16 & 1024/24/16 \\
Global batch & 256 & 128 & 256 & 256 & 160 & 512 \\
Base learning rate & \(5{\times}10^{-4}\) & \(10^{-4}\) & \(10^{-4}\) & \(10^{-4}\) & \(2{\times}10^{-4}\) & \(10^{-4}\) / \(10^{-5}\)$^\S$ \\
Vision LR multiplier & 0.1 & 0.1 & 0.05 & 0.05 & 0.1 & 0.01 \\
Training iterations & 200k & 200k & 100k & 200k & 200k & 200k$^\S$ \\
Precision & \multicolumn{6}{c}{bfloat16 activation autocast; FP32 trainable parameter storage} \\
Language model & \multicolumn{6}{c}{None: fixed codes + random action-trained lookup} \\
\bottomrule
\end{tabular}}
\end{table}

\subsubsection[Pretrained Text Optimization]{\rev{Pretrained Text Optimization}}
\label{app:text_optimization}
\rev{The original component ablations use AdamW with \((\beta_1,\beta_2)=(0.9,0.95)\) and zero weight decay. Only the action heads are trained during the first 1,000 updates. Thereafter, the BERT and SigLIP2 embedding tables and the full 24-layer SigLIP2 text tower are fine-tuned under the action objective. The tower's pooled projection head is discarded. The final-layer, 1,024-dimensional contextual token states enter the joint action Transformer without an additional text adapter.}

\noindent\begin{minipage}{\linewidth}
\centering
\small
\setlength{\tabcolsep}{4pt}
\captionsetup{font=small,skip=3pt}
\captionof{table}{\textbf{LIBERO default-recipe seed variance.} Success (\%), 500 episodes per suite.}
\label{tab:seed_variance_libero}
\begin{tabular}{c|ccccc}
\toprule
Seed & Spatial & Object & Goal & Long & Avg \\
\midrule
0 & 98.6 & 99.0 & 96.4 & 91.4 & 96.4 \\
1 & 98.4 & 99.6 & 97.4 & 92.2 & 96.9 \\
2 & 98.2 & 98.6 & 96.0 & 91.8 & 96.2 \\
\midrule
Mean $\pm$ std & 98.4 $\pm$ 0.2 & 99.1 $\pm$ 0.4 & 96.6 $\pm$ 0.6 & 91.8 $\pm$ 0.3 & \textbf{96.5 $\pm$ 0.3} \\
\bottomrule
\end{tabular}
\par\vspace{5pt}

\captionof{table}{\textbf{WidowX default-recipe seed variance.} Success (\%), 24 episodes per task.}
\label{tab:seed_variance_widowx}
\begin{tabular}{c|ccccc}
\toprule
Seed & Stack & Eggplant & Carrot & Spoon & Avg \\
\midrule
0 & 79.2 & 87.5 & 100.0 & 100.0 & 91.7 \\
1 & 79.2 & 83.3 & 100.0 & 100.0 & 90.6 \\
2 & 91.7 & 87.5 & 100.0 & 100.0 & 94.8 \\
\midrule
Mean $\pm$ std & 83.3 $\pm$ 5.9 & 86.1 $\pm$ 2.0 & 100.0 $\pm$ 0.0 & 100.0 $\pm$ 0.0 & \textbf{92.4 $\pm$ 1.8} \\
\bottomrule
\end{tabular}
\par\vspace{5pt}

\captionof{table}{\rev{\textbf{CALVIN ABC$\rightarrow$D default-recipe seed variance.} Success rates are percentages.}}
\label{tab:seed_variance_calvin}
\begin{tabular}{l*{6}{r}}
\toprule
\textbf{Seed} & \textbf{SR@1} & \textbf{SR@2} & \textbf{SR@3} & \textbf{SR@4} & \textbf{SR@5} & \textbf{Avg. len.} \\
\midrule
0 & 99.4 & 93.0 & 85.8 & 78.6 & 67.6 & 4.24 \\
1 & 98.0 & 91.8 & 84.0 & 77.0 & 68.6 & 4.19 \\
2 & 98.8 & 93.4 & 84.4 & 80.0 & 71.8 & 4.28 \\
\midrule
Mean $\pm$ std & 98.7 $\pm$ 0.6 & 92.7 $\pm$ 0.7 & 84.7 $\pm$ 0.8 & 78.5 $\pm$ 1.2 & 69.3 $\pm$ 1.8 & 4.24 $\pm$ 0.04 \\
\bottomrule
\end{tabular}
\end{minipage}

\rev{The text learning rate is \(\mathrm{lr}_{\mathrm{text}}=\mathrm{lr}_{\mathrm{base}}\times\mathrm{coef}_{\mathrm{text}}\), using the benchmark-specific base rates in \Cref{tab:hparams}. The original coefficient is 1, following the default random-embedding configuration. We independently sweep \(\{1.0,0.1,0.05,0.0\}\) for each pretrained variant on LIBERO, WidowX, and CALVIN, holding the remaining settings fixed within each sweep. A coefficient of zero freezes the text component. \Cref{app:extra_text_results} gives the selected coefficients and corresponding scores.}

\subsection{Flow-Matching Instantiation of the Action Head}
\label{app:head}
Every reported run instantiates the head of \Cref{sec:method} with conditional flow matching~\citep{lipman2022flow}.
For a normalized expert chunk \(x\) and \(\epsilon\sim\mathcal N(0,I)\), we sample \(\widetilde{\tau}\sim\operatorname{Beta}(1.5,1)\), set \(\tau=0.999\widetilde{\tau}+0.001\), and construct
\[
x_\tau=\tau\epsilon+(1-\tau)x,
\qquad u_\tau=\epsilon-x,
\]
so that the auxiliary input is \(z=(\tau,x_\tau)\) and the target is \(y=u_\tau\).
The training objective is
\[
\mathcal L_{\mathrm{FM}}
=\mathbb E_{x,\epsilon,\tau}
\left[
\left\|v_\theta(x_\tau,c_t,\tau)-u_\tau\right\|_2^2
\right].
\]
Inference initializes \(x_1\sim\mathcal N(0,I)\) and applies explicit Euler integration from \(\tau=1\) to \(0\) using the \(K\) steps recorded in \Cref{tab:interfaces}.

\begin{figure}[t]
\centering
\includegraphics[width=1.0\linewidth]{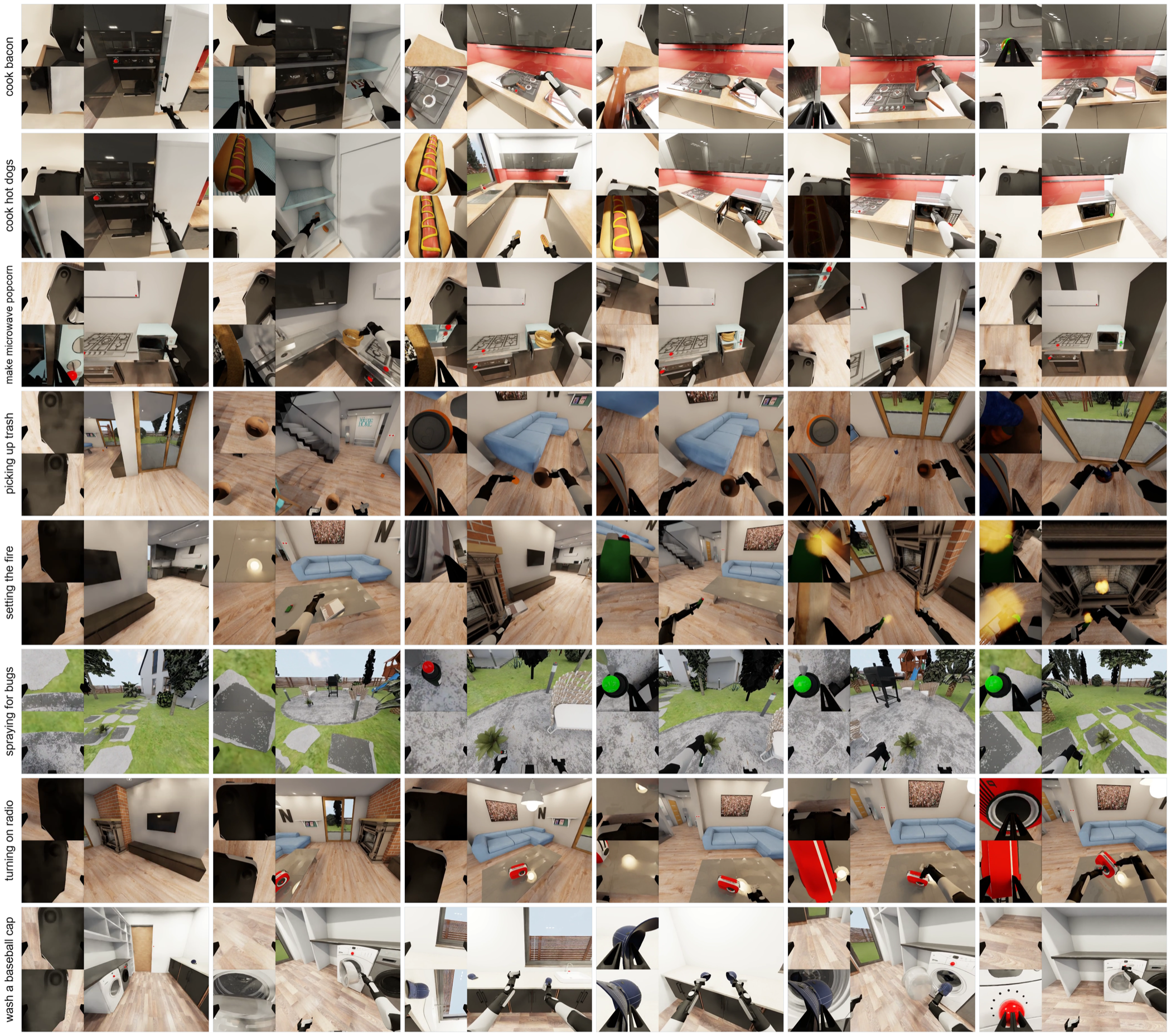}
\caption{\textbf{BEHAVIOR-1K rollouts.}
Eight household activities, one goal-satisfying episode each (\(q{=}1\)), with six frames sampled across the episode from left to right.
Each frame shows the three onboard views the policy receives: egocentric, left wrist, and right wrist.
Episode lengths differ by an order of magnitude across these rows, from roughly 2{,}100 simulation steps for \emph{turning on radio} to 14{,}700 for \emph{setting the fire}.}
\label{fig:b1k_rollouts}
\end{figure}

\subsection{Parameter Accounting}
\begin{table}[h]
\caption{\textbf{Approximate architectural parameter accounting for the LM-free SigLIP2 models.}
The table separates the vision backbone, action-trained instruction embedding, action Transformer, and pretrained language-side parameters.}
\label{tab:params}
\centering
\scriptsize
\setlength{\tabcolsep}{10pt}
\begin{tabular}{lrr}
\toprule
Component & Manipulation default  \\
\midrule
SigLIP2 vision model & \(\sim303\)M \\
Random instruction embedding & \(\sim31\)M \\
Action Transformer & \(\sim303\)M  \\
Total incl.\ projections/heads & \(\sim639\)M  \\
Pretrained language-side parameters & \textbf{0}  \\
\bottomrule
\end{tabular}
\end{table}

\method{} is not a small policy in total parameter count.
Its approximately 31M instruction-embedding parameters are random at initialization and action-trained, not pretrained language parameters.
With FP32 parameters, gradients, and two Adam moments, the 31M instruction-embedding table alone occupies approximately 0.5 decimal GB of training state before framework and sharding overhead.

\subsection{BEHAVIOR-1K Qualitative Rollouts}
\label{app:rollouts}
\Cref{fig:b1k_rollouts} shows the executed behavior on eight such activities.

% \clearpage
\subsection{Seed Variance: A Noise Floor for the Ablations}
\label{app:seed_variance}
\rev{We train the default configuration with seeds 0, 1, and 2 on LIBERO, WidowX, and CALVIN using the protocols in \Cref{tab:protocols} (500 episodes per LIBERO suite and 24 per WidowX task). We report means and standard deviations across seeds. \Cref{tab:seed_variance_libero,tab:seed_variance_widowx,tab:seed_variance_calvin} gives per-seed scores, including the seed-0 results used for headline comparisons.}

LIBERO is tight (range \(0.7\) points, std \(0.3\)); WidowX is far noisier (range \(4.2\) points, std \(1.8\)), driven almost entirely by the Stack task.

\rev{We repeat the text-component ablations on LIBERO and WidowX with seeds 0, 1, and 2 under the original ablation configurations.}

\rev{BERT embeddings remain close to the default on LIBERO. On WidowX Eggplant, the two SigLIP2 text variants score below both the default and BERT in the three-seed averages. The variants use a vocabulary of roughly 256k SentencePiece entries instead of 30,522 WordPiece entries and change the segmentation of \textit{eggplant}. They also increase the table from roughly 31M to 262M parameters. These changes and the representation itself may affect how supervision transfers across instructions. }

\noindent\begin{minipage}{\linewidth}
\centering
\small
\setlength{\tabcolsep}{4pt}
\captionsetup{font=small,skip=3pt}
\captionof{table}{\rev{\textbf{Three-seed text-component ablations on LIBERO.} Success in percent, mean and standard deviation.}}
\label{tab:seed_text_libero}
\begin{tabular}{l*{5}{r}}
\toprule
\textbf{Configuration} & \textbf{Spatial} & \textbf{Object} & \textbf{Goal} & \textbf{Long} & \textbf{Avg.} \\
\midrule
Default & 98.4 $\pm$ 0.2 & 99.1 $\pm$ 0.4 & 96.6 $\pm$ 0.6 & 91.8 $\pm$ 0.3 & 96.5 $\pm$ 0.3 \\
BERT embedding table & 99.6 $\pm$ 0.2 & 99.6 $\pm$ 0.4 & 96.7 $\pm$ 0.6 & 88.3 $\pm$ 2.3 & 96.1 $\pm$ 0.7 \\
SigLIP2 embedding table & 97.8 $\pm$ 0.8 & 98.1 $\pm$ 1.7 & 94.1 $\pm$ 1.7 & 85.3 $\pm$ 3.2 & 93.8 $\pm$ 0.7 \\
SigLIP2 text encoder & 96.5 $\pm$ 0.2 & 98.9 $\pm$ 0.7 & 85.5 $\pm$ 2.1 & 84.6 $\pm$ 2.6 & 91.4 $\pm$ 0.5 \\
\bottomrule
\end{tabular}
\par\vspace{5pt}

\captionof{table}{\rev{\textbf{Three-seed text-component ablations on WidowX.} Success in percent, mean and standard deviation.}}
\label{tab:seed_text_widowx}
\begin{tabular}{l*{5}{r}}
\toprule
\textbf{Configuration} & \textbf{Carrot} & \textbf{Eggplant} & \textbf{Spoon} & \textbf{Stack} & \textbf{Avg.} \\
\midrule
Default & 100.0 $\pm$ 0.0 & 86.1 $\pm$ 2.0 & 100.0 $\pm$ 0.0 & 83.3 $\pm$ 5.9 & 92.4 $\pm$ 1.8 \\
BERT embedding table & 97.2 $\pm$ 2.0 & 88.9 $\pm$ 4.0 & 93.1 $\pm$ 1.9 & 73.6 $\pm$ 2.0 & 88.2 $\pm$ 1.7 \\
SigLIP2 embedding table & 97.2 $\pm$ 2.0 & 54.2 $\pm$ 15.6 & 100.0 $\pm$ 0.0 & 83.3 $\pm$ 6.8 & 83.7 $\pm$ 6.0 \\
SigLIP2 text encoder & 93.1 $\pm$ 1.9 & 44.4 $\pm$ 10.4 & 100.0 $\pm$ 0.0 & 84.7 $\pm$ 8.5 & 80.6 $\pm$ 4.7 \\
\bottomrule
\end{tabular}
\end{minipage}

\subsection{Further Discussion}

\paragraph{Reproducibility Statement}
\Cref{sec:setup} summarizes the benchmark suite and evaluation setup.
The appendix provides evaluation protocols, benchmark-specific hyperparameters, control interfaces, and flow-matching implementation details (\Cref{tab:protocols}; \Cref{tab:hparams}; \Cref{tab:interfaces}; \Cref{app:head}).
Code is available at \url{https://frontierrobo.github.io/SimVLA2}.

\paragraph{Ethics Statement}
The reported benchmarks are simulated or benchmark-mediated robot-control evaluations and contain no human-subject data.
Benchmark results do not establish safe deployment; physical robot systems require collision handling, uncertainty monitoring, workspace limits, hardware interlocks, and human oversight.

% \newpage
% \clearpage
\section[Additional Experiments and Diagnostics]{\rev{Additional Experiments and Diagnostics}}
\label{app:rebuttal_results}

\subsection[System-Level Comparison and Language-Pathway Replacement]{\rev{System-Level Comparison and Language-Pathway Replacement}}
\label{app:simvla_removal}
\rev{SimVLA and \method{} use the same candidate values for shared hyperparameters: action horizon \(H\in\{10,20,30\}\), base learning rate \(\{5{\times}10^{-5},10^{-4},2{\times}10^{-4},5{\times}10^{-4}\}\), warm-up steps \(\{0,1000\}\), batch size \(\{128,256\}\), and cosine decay or no scheduler. The search also covers data shuffling and action/proprioception normalization, with at most 200k steps. The reported \method{} configurations are in \Cref{tab:interfaces,tab:hparams}.}

\rev{On Google Robot we retain SimVLA's pretrained vision tower, visual connector, and action Transformer. We remove the language model and its output head, pass visual features directly to the action Transformer, and supply randomly initialized, action-trained WordPiece embeddings. The two variants share training data, optimization settings, training budget, action chunking, and evaluation protocol. The ablation measures replacement of the complete pretrained language pathway. The average decreases by 1.6 percentage points.}

\begin{table}[!htbp]
\centering
\caption{\rev{\textbf{SimVLA language-pathway replacement on Google Robot.} Success rates in percent.}}
\label{tab:simvla_removal}
\small
\resizebox{\ifdim\width>\linewidth\linewidth\else\width\fi}{!}{%
\begin{tabular}{l*{4}{r}}
\toprule
\textbf{Configuration} & \textbf{Pick} & \textbf{Move} & \textbf{Open} & \textbf{Avg.} \\
\midrule
SimVLA & 82.3 & 81.0 & 67.7 & 77.0 \\
SimVLA with action-trained token embeddings & 80.7 & 77.3 & 68.3 & 75.4 \\
\bottomrule
\end{tabular}}
\end{table}

% \clearpage
\subsection[Instruction Rewording and Training-Token Coverage]{\rev{Instruction Rewording and Training-Token Coverage}}
\label{app:instruction_diagnostics}
\rev{We evaluate the default and pretrained text variants on LIBERO-Para~\citep{LIBEROPara}, which rewords instructions for the underlying LIBERO-Goal tasks. The compositional category denotes combined linguistic changes. A \emph{training-unseen} token is a content-token ID absent from the action-training instruction vocabulary. Both WordPiece and SigLIP2 SentencePiece tokenize these prompts without generating tokenizer-level \texttt{[UNK]} tokens. The default's unseen token IDs retain their individual initial vectors under the zero-weight-decay setup. The LIBERO training vocabulary contains 112 distinct strings and 87 distinct content WordPiece IDs.}

\rev{The default attains \(60.12\%\) overall. Its success is approximately \(80\%\) when up to one quarter of the content tokens are training-unseen and \(34.14\%\) above one half. Within the shared WordPiece buckets, BERT scores higher on prompts with complete coverage, and the default scores higher in the three groups containing unseen tokens. The association with token coverage bounds the observed tolerance to rewording.}

\begin{table}[!htbp]
\centering
\caption{\rev{\textbf{LIBERO-Para comparison with published scores.} Goal and paraphrase success rates are percentages.}}
\label{tab:para_comparison}
\small
\resizebox{\ifdim\width>\linewidth\linewidth\else\width\fi}{!}{%
\begin{tabular}{l*{6}{r}}
\toprule
\textbf{Method} & \textbf{Goal} & \textbf{Para} & \textbf{Action} & \textbf{Object} & \textbf{Compositional} & \textbf{PRIDE ($\alpha=0.5$)} \\
\midrule
Xiaomi-Robotics-0 & 98.8 & 76.0 & 89.6 & 88.6 & 70.9 & 69.2 \\
$\pi_{0.5}$ & 97.6 & 71.4 & 89.9 & 87.8 & 64.6 & 65.4 \\
OpenVLA-OFT (goal / mixed) & 97.9 / 96.1 & 64.7 / 63.7 & 89.1 / 89.5 & 74.2 / 71.4 & 56.7 / 55.5 & 58.8 / 56.3 \\
X-VLA & 97.8 & 62.1 & 83.0 & 84.9 & 54.0 & 52.7 \\
VLA-Adapter & 98.2 & 46.3 & 73.7 & 75.0 & 35.7 & 36.1 \\
$\pi_{0.5}$ (expert-only, frozen VLM) & 78.6 & 39.1 & 58.9 & 74.9 & 30.2 & 32.0 \\
SimVLA 2.0 & 96.4 & 60.12 & 82.53 & 74.52 & 52.28 & 49.82 \\
\bottomrule
\end{tabular}}
\end{table}

\begin{table}[!htbp]
\centering
\caption{\rev{\textbf{LIBERO-Para text-component comparison.} Training-unseen token rates and success rates are percentages.}}
\label{tab:para_variants}
\small
\resizebox{\ifdim\width>\linewidth\linewidth\else\width\fi}{!}{%
\begin{tabular}{l*{5}{r}}
\toprule
\textbf{Text component} & \textbf{Training-unseen} & \textbf{Para SR} & \textbf{Action} & \textbf{Object} & \textbf{Compositional} \\
\midrule
Default random table & 39.94\% & 60.12 & 82.53 & 74.52 & 52.28 \\
BERT embedding table & 39.94\% & 47.75 & 67.47 & 63.71 & 40.57 \\
SigLIP2 embedding table & 41.34\% & 44.11 & 55.17 & 77.61 & 37.93 \\
SigLIP2 text encoder & 41.34\% & 42.25 & 58.05 & 52.51 & 36.72 \\
\bottomrule
\end{tabular}}
\end{table}

\begin{table}[!htbp]
\centering
\caption{\rev{\textbf{Training-token coverage across instruction evaluations.}}}
\label{tab:token_coverage}
\small
\resizebox{\ifdim\width>\linewidth\linewidth\else\width\fi}{!}{%
\begin{tabular}{l*{2}{r}}
\toprule
\textbf{Evaluation} & \textbf{Unseen tokens} & \textbf{Coverage details} \\
\midrule
LIBERO-Para, WordPiece & 39.94\% & 98.63\% of prompts contain at least one unseen token \\
LIBERO-Para, SentencePiece & 41.34\% & 99.14\% of prompts contain at least one unseen token \\
LIBERO-PRO Sem candidate set, WordPiece & 15.51\% & 157/1,012 token occurrences across 120 candidate paraphrases \\
SimplerEnv Google Robot, WordPiece & 0\% & Evaluation token IDs are covered by the training vocabulary \\
\bottomrule
\end{tabular}}
\end{table}

\begin{table}[!htbp]
\centering
\caption{\rev{\textbf{LIBERO-Para success by training-unseen content-token share.} Success in percent. Buckets use each tokenizer separately.}}
\label{tab:para_coverage_bins}
\small
\resizebox{\ifdim\width>\linewidth\linewidth\else\width\fi}{!}{%
\begin{tabular}{l*{4}{r}}
\toprule
\textbf{Unseen-token share} & \textbf{Default} & \textbf{BERT table} & \textbf{SigLIP2 table} & \textbf{SigLIP2 encoder} \\
\midrule
0\% & 80.36 & 89.29 & 71.43 & 65.71 \\
(0\%, 25\%] & 79.47 & 76.83 & 67.56 & 55.85 \\
(25\%, 50\%] & 62.12 & 46.21 & 41.97 & 41.52 \\
\(>50\%\) & 34.14 & 20.00 & 30.65 & 32.83 \\
Overall & 60.12 & 47.75 & 44.11 & 42.25 \\
\bottomrule
\end{tabular}}
\end{table}

\clearpage
\subsection[Whole-Instruction Task-ID Baseline]{\rev{Whole-Instruction Task-ID Baseline}}
\label{app:whole_instruction}
\rev{We replace the instruction token sequence with one learned 1,024-dimensional embedding per registered training instruction string, retaining the vision encoder and action Transformer. Unseen full strings map to a shared fallback embedding. This baseline tests exact-string lookup. On standard LIBERO, whole-instruction lookup scores close to the token-level policy. The token-level policy is stronger under instruction rewording. On CALVIN, all 34 evaluation instruction strings are absent verbatim from the training inventory. These tests leave open how the token-level policy uses familiar lexical cues.}

\begin{table}[!htbp]
\centering
\caption{\rev{\textbf{Whole-instruction lookup across evaluations.}}}
\label{tab:task_id_summary}
\small
\resizebox{\ifdim\width>\linewidth\linewidth\else\width\fi}{!}{%
\begin{tabular}{l*{2}{r}}
\toprule
\textbf{Evaluation} & \textbf{Token-level} & \textbf{Whole-instruction} \\
\midrule
Standard LIBERO, average success (\%) & 96.4 & 94.8 \\
LIBERO-Para, success (\%) & 60.12 & 5.74 \\
LIBERO-PRO Sem, success (\%) & 96.5 & 30.5 \\
CALVIN ABC$\rightarrow$D, average chain length & 4.24 & 0.03 \\
WidowX, average success (\%) & 91.7 & 50.0 \\
WidowX Eggplant, success (\%) & 87.5 & 4.17 \\
\bottomrule
\end{tabular}}
\end{table}

\begin{table}[!htbp]
\centering
\caption{\rev{\textbf{Whole-instruction lookup on the LIBERO suites.} Success in percent.}}
\label{tab:task_id_libero}
\small
\resizebox{\ifdim\width>\linewidth\linewidth\else\width\fi}{!}{%
\begin{tabular}{l*{6}{r}}
\toprule
\textbf{Method} & \textbf{Spatial} & \textbf{Object} & \textbf{Goal} & \textbf{Long} & \textbf{Avg.} & \textbf{LIBERO-Para} \\
\midrule
SimVLA 2.0 (Default) & 98.6 & 99.0 & 96.4 & 91.4 & 96.4 & 60.12 \\
Whole-Instruction Task-ID & 97.6 & 99.4 & 96.2 & 86.0 & 94.8 & 5.74 \\
\bottomrule
\end{tabular}}
\end{table}

\begin{table}[!htbp]
\centering
\caption{\rev{\textbf{Whole-instruction lookup on CALVIN ABC$\rightarrow$D.}}}
\label{tab:task_id_calvin}
\small
\resizebox{\ifdim\width>\linewidth\linewidth\else\width\fi}{!}{%
\begin{tabular}{l*{6}{r}}
\toprule
\textbf{Method} & \textbf{SR@1 (\%)} & \textbf{SR@2 (\%)} & \textbf{SR@3 (\%)} & \textbf{SR@4 (\%)} & \textbf{SR@5 (\%)} & \textbf{Avg. len.} \\
\midrule
SimVLA 2.0 (Default) & 99.4 & 93.0 & 85.8 & 78.6 & 67.6 & 4.24 \\
Whole-Instruction Task-ID & 3.0 & 0.1 & 0.0 & 0.0 & 0.0 & 0.03 \\
\bottomrule
\end{tabular}}
\end{table}

\begin{table}[!htbp]
\centering
\caption{\rev{\textbf{Whole-instruction lookup by WidowX task.} Success in percent.}}
\label{tab:task_id_widowx}
\small
\resizebox{\ifdim\width>\linewidth\linewidth\else\width\fi}{!}{%
\begin{tabular}{l*{5}{r}}
\toprule
\textbf{Method} & \textbf{Carrot} & \textbf{Eggplant} & \textbf{Spoon} & \textbf{Stack} & \textbf{Avg.} \\
\midrule
SimVLA 2.0 (Default) & 100.0 & 87.5 & 100.0 & 79.2 & 91.7 \\
Whole-Instruction Task-ID & 91.7 & 4.17 & 79.2 & 25.0 & 50.0 \\
\bottomrule
\end{tabular}}
\end{table}

\clearpage
\subsection[Text Learning-Rate Sensitivity and CALVIN Component Ablations]{\rev{Text Learning-Rate Sensitivity and CALVIN Component Ablations}}
\label{app:extra_text_results}
\Cref{app:text_optimization} describes the sweep. The coefficients below identify the selected settings. On LIBERO, the SigLIP2 text tower reaches \(96.3\%\) at a coefficient of \(0.1\), compared with \(91.3\%\) in the original setting. BERT embeddings achieve \(96.8\%\), compared with \(96.4\%\) for the default. The evaluated pretrained text components offer no consistent advantage across the benchmark settings. CALVIN also shows substantial sensitivity to instruction conditioning and visual pretraining.

\begin{table}[!htbp]
\centering
\caption{\rev{\textbf{Selected text learning rates on LIBERO.} Success rates in percent.}}
\label{tab:text_lr_libero}
\small
\resizebox{\ifdim\width>\linewidth\linewidth\else\width\fi}{!}{%
\begin{tabular}{l*{6}{r}}
\toprule
\textbf{Text component} & \textbf{Coefficient} & \textbf{Spatial} & \textbf{Object} & \textbf{Goal} & \textbf{Long} & \textbf{Avg.} \\
\midrule
Random table & Default & 98.6 & 99.0 & 96.4 & 91.4 & 96.4 \\
BERT emb. table & 1.0 & 99.8 & 99.0 & 97.6 & 90.8 & 96.8 \\
SigLIP2 emb. table & 1.0 & 98.8 & 99.2 & 96.4 & 84.4 & 94.7 \\
SigLIP2 text encoder & 0.1 & 98.8 & 99.6 & 96.6 & 90.2 & 96.3 \\
\bottomrule
\end{tabular}}
\end{table}

\begin{table}[!htbp]
\centering
\caption{\rev{\textbf{Selected text learning rates on WidowX.} Success rates in percent.}}
\label{tab:text_lr_widowx}
\small
\resizebox{\ifdim\width>\linewidth\linewidth\else\width\fi}{!}{%
\begin{tabular}{l*{6}{r}}
\toprule
\textbf{Text component} & \textbf{Coefficient} & \textbf{Carrot} & \textbf{Eggplant} & \textbf{Spoon} & \textbf{Stack} & \textbf{Avg.} \\
\midrule
Random table & Default & 100.0 & 87.5 & 100.0 & 79.2 & 91.7 \\
BERT emb. table & 0.05 & 100.0 & 91.7 & 95.8 & 75.0 & 90.6 \\
SigLIP2 emb. table & 0.05 & 95.8 & 50.0 & 100.0 & 83.3 & 82.3 \\
SigLIP2 text encoder & 0.05 & 95.8 & 41.7 & 100.0 & 83.3 & 80.2 \\
\bottomrule
\end{tabular}}
\end{table}

\begin{table}[!htbp]
\centering
\caption{\rev{\textbf{Text learning rates and additional CALVIN ABC$\rightarrow$D ablations.} Success rates are percentages. Dashes indicate non-applicable coefficients.}}
\label{tab:text_lr_calvin}
\small
\resizebox{\ifdim\width>\linewidth\linewidth\else\width\fi}{!}{%
\begin{tabular}{l*{7}{r}}
\toprule
\textbf{Configuration} & \textbf{Coefficient} & \textbf{SR@1} & \textbf{SR@2} & \textbf{SR@3} & \textbf{SR@4} & \textbf{SR@5} & \textbf{Avg. len.} \\
\midrule
Random table & Default & 99.4 & 93.0 & 85.8 & 78.6 & 67.6 & 4.24 \\
BERT emb. table & 1.0 & 91.3 & 78.9 & 64.6 & 54.2 & 43.6 & 3.33 \\
SigLIP2 emb. table & 1.0 & 97.6 & 90.5 & 81.5 & 73.3 & 64.4 & 4.07 \\
SigLIP2 text encoder & 0.1 & 96.4 & 90.9 & 82.6 & 75.5 & 67.1 & 4.13 \\
\midrule
No instruction & --- & 34.2 & 7.7 & 2.3 & 0.6 & 0.0 & 0.45 \\
Whole-instruction Task-ID & 1.0 & 3.0 & 0.1 & 0.0 & 0.0 & 0.0 & 0.03 \\
DINOv3 vision backbone & 1.0 & 82.4 & 60.8 & 44.4 & 31.4 & 20.0 & 2.39 \\
Randomly initialized SigLIP2 vision tower & 1.0 & 55.1 & 27.2 & 12.0 & 4.1 & 1.8 & 1.00 \\
\bottomrule
\end{tabular}}
\end{table}

\end{document}